\documentclass[10pt,twocolumn,letterpaper]{article}

\usepackage{iccv}
\usepackage{times}
\usepackage{epsfig}
\usepackage{graphicx}
\usepackage{amsmath}
\usepackage{amssymb}

\usepackage{booktabs}
\usepackage{enumerate}
\usepackage{float}
\usepackage{multirow}
\usepackage{bbding}
\usepackage{makecell}

\usepackage[accsupp]{axessibility}

\usepackage{subcaption}
\usepackage[pagebackref=true,breaklinks=true,letterpaper=true,colorlinks,bookmarks=false]{hyperref}

\usepackage[capitalize]{cleveref}
\crefname{section}{Sec.}{Secs.}
\Crefname{section}{Section}{Sections}
\Crefname{table}{Table}{Tables}
\crefname{table}{Tab.}{Tabs.}

\iccvfinalcopy 

\def\iccvPaperID{3035} 

\ificcvfinal\fi

\def\mywork{U2MOT}

\newcommand{\lk}[1]{#1}
\newcommand{\liuk}[1]{#1}

\newcommand{\minus}{\text{-}}
\newcommand{\topa}[1]{\underline{\textbf{#1}}}
\newcommand{\topb}[1]{#1}
\newcommand{\mappendix}[1]{Appendix \textcolor{red}{#1}}

\begin{document}

\title{Uncertainty-aware Unsupervised Multi-Object Tracking}

\author{
Kai Liu$^{1}$\footnotemark[1], $\;$ Sheng Jin$^{2}$, $\;$ Zhihang Fu$^{2}$\footnotemark[2], $\;$ Ze Chen$^{2}$, $\;$ Rongxin Jiang$^{1}$, $\;$ Jieping Ye$^{2}$
\vspace{0.5em}\\
$^{1}$Zhejiang University, $\;$ $^{2}$Alibaba DAMO Academy\\
}

\maketitle
\ificcvfinal\thispagestyle{empty}\fi

\renewcommand{\thefootnote}{\fnsymbol{footnote}}
\footnotetext[1]{This work was done when Kai Liu worked as a research intern at Alibaba DAMO Academy. Email: kail@zju.edu.cn.}
\footnotetext[2]{Corresponding author. Email: zhihang.fzh@alibaba-inc.com.}

\begin{abstract}
    Without manually annotated identities, unsupervised multi-object trackers are inferior to learning reliable feature embeddings.
    It causes the similarity-based inter-frame association stage also be error-prone, where an \textbf{uncertainty} problem arises.
    The frame-by-frame accumulated uncertainty prevents trackers from learning the consistent feature embedding against time variation.
    To avoid this uncertainty problem, recent self-supervised techniques are adopted, whereas they failed to capture temporal relations. The inter-frame uncertainty still exists.
    In fact, this paper argues that though the uncertainty problem is inevitable, it is possible to leverage the uncertainty itself to improve the learned consistency in turn.
    Specifically, an uncertainty-based metric is developed to verify and rectify the risky associations. The resulting accurate pseudo-tracklets boost learning the feature consistency.
    And accurate tracklets can incorporate temporal information into spatial transformation. This paper proposes a tracklet-guided augmentation strategy to simulate the tracklet's motion,  which adopts a hierarchical uncertainty-based sampling mechanism for hard sample mining.
    The ultimate unsupervised MOT framework, namely~\mywork, is proven effective on MOT-Challenges and VisDrone-MOT benchmark. 
    \mywork~
    achieves a SOTA performance among the published supervised and unsupervised trackers.
    Code is available at \url{https://github.com/alibaba/u2mot/}.
\end{abstract}


\section{Introduction}
\label{sec:intro}

\begin{figure}[t]
  \centering
  \includegraphics[width=\linewidth]{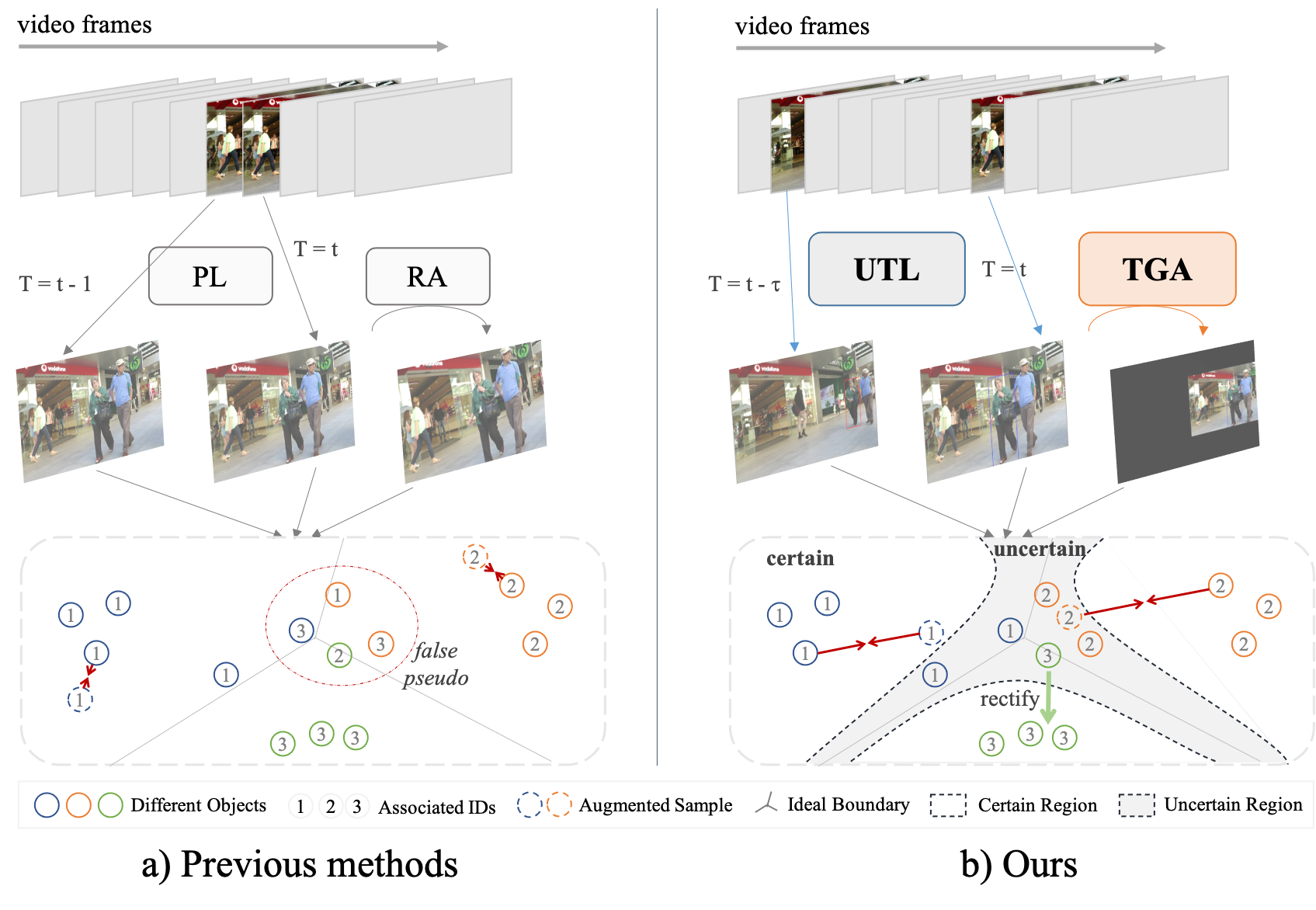}
  \caption{\textbf{Method Comparison.} a) Previous methods utilize \lk{pseudo-labeled (PL)} adjacent frames or \lk{randomly augmented (RA)} samples. b) Our method adopts \lk{uncertainty-aware tracklet-labeling (UTL)} to maintain \liuk{consistent pseudo-tracklets}, and exploits tracklet-guided augmentation \lk{(TGA)} to improve the embedding consistency }
  \label{fig:intro_moti}
\end{figure}

Multi-object tracking (MOT)~\cite{milan2016mot16,bewley2016simple,wojke2017simple} has been widely deployed in real-world applications, including surveillance analysis~\cite{oh2011large,zhu2018vision}, autonomous driving~\cite{geiger2012we,sun2020scalability}, intelligent robots~\cite{said2012real,bescos2021dynaslam}, \etc. 
\liuk{The goal of MOT task is to detect all target objects and simultaneously keep their respective feature embeddings \textbf{consistent}, regardless of the change of their shapes and angles over a period of time~\cite{wojke2017simple,zhang2021fairmot}.
However, the core issue of unsupervised MOT task is lacking the annotated ID-supervision to confirm the consistency of a certain target, especially when its shape and angle are varied over time~\cite{karthik2020simple,wang2020cycas,liu2022online,li2022unsupervised,shuai2022id}. 
When training an unsupervised tracker, since the learned feature embedding is unreliable, the similarity-based association stage is error-prone. Propagating pseudo identities frame-by-frame leads to uncertainty in the resulting pseudo-tracklets, which accumulates in per-frame associations. This prevents trackers from learning a consistent feature embedding~\cite{wang2020cycas,liu2022online}.
}

\lk{To avoid this problem, self-supervised techniques~\cite{liu2022online,shuai2022id,zhan2021spatial} are utilized to generate augmented samples with perfectly-accurate identities. 
However, \liuk{these commonly-used methods merely take a single frame for augmentation,} while the inter-frame temporal relation is totally ignored. \liuk{It usually leads to sub-optimal performance~\cite{zhang2021fairmot}. The uncertainty problem seems inevitable but remains under-explored.}
In fact, we argue that the uncertainty can be leveraged to maintain consistency in turn, as shown in~\cref{fig:intro_moti}.}

First, uncertainty can guide the construction of pseudo-tracklets.
When the similarity-based inter-frame object association is inaccurate, we propose to introduce a quantified uncertainty measure to find out the possibly wrong associations and further re-associate them. 
Specifically, we find that the association mismatching is accompanied by two phenomenons, including a small similarity margin (similar appearance \etc) and low confidence (object occlusion \etc).
Inspired by these findings, an \textbf{association-uncertainty} metric is proposed to filter the uncertain candidate set, which is further rectified using the tracklet appearance and motion clues. 
The proposed mechanism, termed \textbf{Uncertainty-aware Tracklet-Labeling (UTL)}, generates highly-accurate pseudo-tracklets to learn the embedding consistency.
The proposed UTL has two features: (1) it can directly boost the tracking performance during inference as well. (2) it is complementary to existing methods and can be incorporated with consistent performance.

Second, uncertainty can guide the hard sample augmentation.
Trustworthy pseudo-tracklets can incorporate temporal information into sample augmentation, thereby overcoming the key limitation of current augmentation-based methods. 
To this end, a \textbf{Tracklet-Guided Augmentation (TGA)} strategy is developed to simulate the real motion of pseudo-tracklets. 
TGA generates augmentation samples aligned to the highly-uncertain objects in the pseudo-tracklet for hard example mining, as a high association-uncertainty basically indicates the presence of challenging negative examples.
We specifically develop a hierarchical uncertainty-based sampling mechanism to ensure trustworthy pseudo-tracklets and hard sample augmentations.

The ultimate \mywork~framework is evaluated on MOT17~\cite{milan2016mot16}, MOT20~\cite{dendorfer2020mot20}, and the challenging VisDrone-MOT~\cite{zhu2021visdrone} benchmarks.
The experiments show that \mywork~significantly outperforms previous unsupervised methods (\eg, 62.7\% $v.s.$ 58.6\% of HOTA on MOT20), and achieves SOTA (\eg, 64.2\% HOTA on MOT17) among existing unsupervised and supervised trackers.
\lk{Extensive ablation studies demonstrate the effectiveness of leveraging uncertainty in improving the consistency in turn.}

Contributions of this paper are summarized as follows:
\begin{enumerate}[1)]
    \setlength{\itemsep}{0pt}
    \item \lk{We are the first to leverage uncertainty} in unsupervised multi-object tracking, where an association-level uncertain metric is introduced to verify the pseudo-tracklets, and a hierarchical uncertainty-based sampling mechanism is developed for hard sample generation.  
    \item We propose a novel unsupervised \mywork~framework, where UTL is developed to guarantee the intra-tracklet consistency and TGA is adopted to learn the consistent feature embedding.
    \item \lk{We achieve a SOTA tracking performance among existing methods, and demonstrate the generalized application prospects for the uncertainty metric.} 
\end{enumerate}
\section{Related Work}
\label{sec:related}
\textbf{Pseudo-label-based Unsupervised MOT}.
Existing unsupervised methods generate pseudo-identities in three main ways, including motion-based, cluster-based, and similarity-based methods. In terms of motion-based methods, SimUMOT~\cite{karthik2020simple} adopts SORT~\cite{bewley2016simple} to generate the pseudo-tracklets, which is used to guide the training of re-identification networks. Very recently, UEANet~\cite{li2022unsupervised} uses ByteTrack~\cite{zhang2022bytetrack} to improve the quality of pseudo labels, where ByteTrack excavated the values of low-confident detection boxes. 
However, long-term dependency within pseudo-tracklets is hard to guarantee, and the spatial information is not reliable in irregular camera motions. 
Cluster-based methods ~\cite{fan2018unsupervised,lin2019bottom,shuai2022id} try to iteratively cluster the objects in the whole video to get pseudo-identities. These methods usually lead to sub-optimal performance. A possible reason is that the temporary association within the tracklet is totally ignored.
The similarity-based methods, like Cycas~\cite{wang2020cycas} and OUTrack~\cite{liu2022online}, utilize the cycle-consistency~\cite{wang2019learning} of object similarities between adjacent frames. As time interval extends, the noise of pseudo-label becomes an inconvenient truth. 
Different from existing methods, our \mywork~ designs an uncertainty-based refinement mechanism to obtain accurate associations. Long-term consistency is preserved through identity propagation.





\textbf{Uncertainty Estimation}.
In recent years, uncertainty estimation has been widely explored in classification calibration (\eg, detecting misclassified or out-of-distribution samples) from three main aspects.
Some researchers adopt deterministic networks~\cite{malinin2018predictive,sensoy2018evidential,devries2018learning} or ensembles~\cite{lakshminarayanan2017simple,wen2019batchensemble} to explicitly represent the uncertainty. 
Others adopt the widely-used softmax probability distribution~\cite{hendrycks2016baseline} to evaluate the credibility according to the classification confidence. 
Very recently, the energy model~\cite{liu2020energy,wang2021can} emerges as the widely-exploited metric in the uncertainty estimation, which is theoretically aligned with the probability density of the inputs. 
However, for multi-object tracking, object occlusion and similar appearance always lead to mismatching.
Thus, the uncertain estimation is worth exploring. In this paper, we design an uncertain metric specially for tracklets-based tasks, which is proved effective. 




\textbf{Augmentation Strategy}.
Adaptive augmentation strategies have been extensively studied in image classification~\cite{fawzi2016adaptive,liu2021divaug}, object detection~\cite{wang2019data,ghiasi2021simple}, and representation learning~\cite{bai2022directional,zhang2022rethinking}.
However, random perspective transformation still dominates in unsupervised multi-object tracking~\cite{zhang2021fairmot,shuai2022id}. 
Other researchers present GAN-based augmentation strategies~\cite{jiang2021exploring,zhan2021spatial} for person re-identification. However, these methods fail to generate realistic object tracklets in MOT situations. 
This work integrates the tracklets property into augmentation and focuses on negative hard sample generations, which makes our augmentation strategy task-specific and effective.




\begin{figure*}[t]
  \centering
  \includegraphics[width=1.0\linewidth]{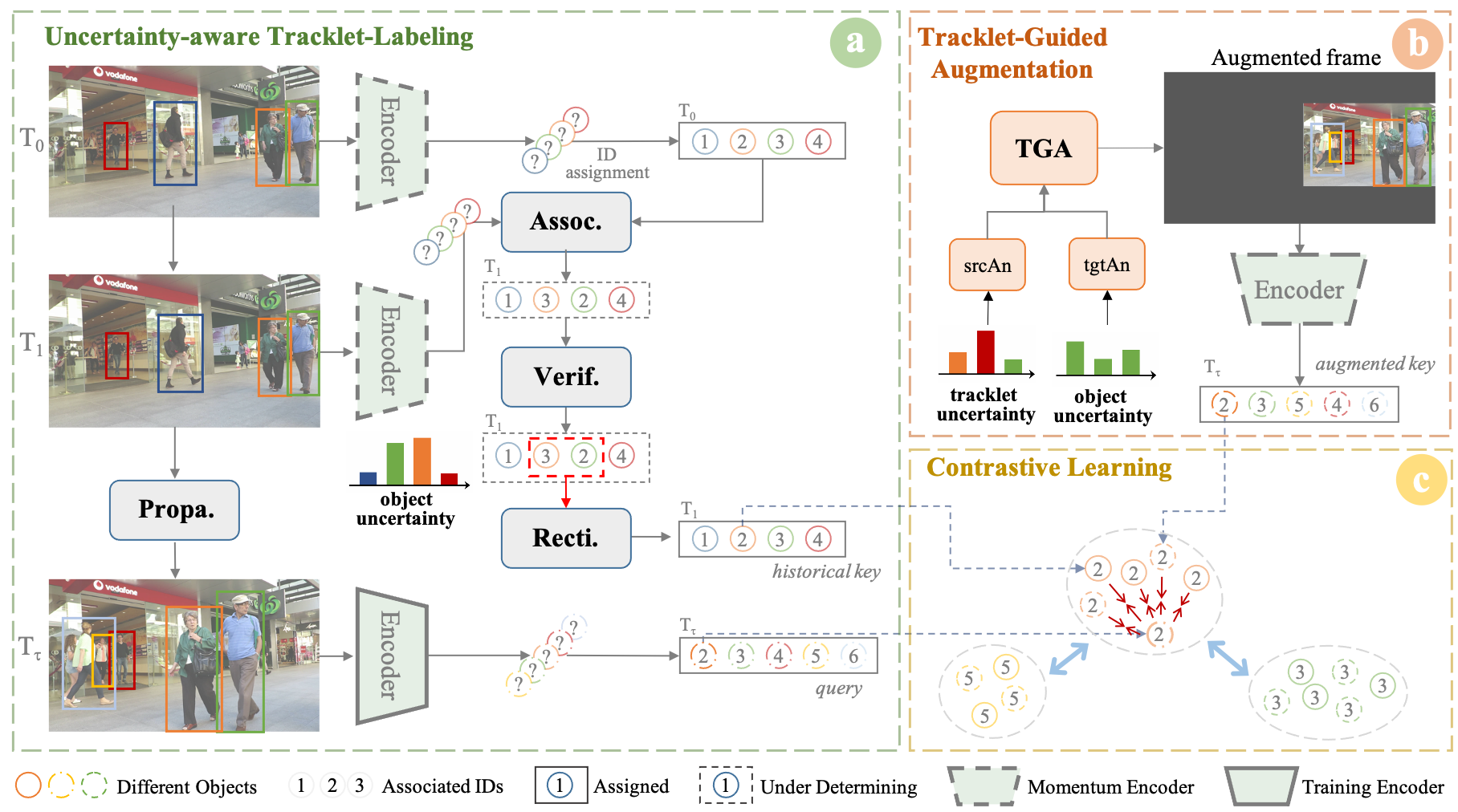}
  \caption{\textbf{Framework of \mywork} (better viewed in color).
  \textbf{a)} We propose an uncertainty metric to verify and rectify the association process. 
  Accurate pseudo-tracklets are propagated frame-by-frame.
  \textbf{b)} Then an anchor pair 
  is selected based on tracklet-level and object-level uncertainties. 
  \lk{Tracklet's motion information is used to guide the augmentation.} \textbf{c)} Contrastive learning is adopted to train the tracker, which pulls the objects within the tracklet together and pushes different tracklets apart.}
  \label{fig:method_fmwk}
\end{figure*}
\section{Methodology}
\label{sec:method}

\subsection{Overview}
\label{sec:method_fmwk}

As shown in~\cref{fig:method_fmwk}, the proposed unsupervised MOT framework is trained with the widely-used contrastive learning technique~\cite{chen2020simple,he2020momentum}. 
\lk{Specifically, for multi-object tracking}, objects within the tracklet ($\boldsymbol{k}_{+}$) should be pulled together and different tracklets ($\boldsymbol{k}_{-}$) should be separated. It can be mathematically formulated as:

\begin{equation}
    \mathcal{L}_{cl}( \boldsymbol{q}; \boldsymbol{k}_{+}; \boldsymbol{k}_{-} )= 
    - \log \frac{\exp(\boldsymbol{q} \cdot \boldsymbol{k}_{+} / \epsilon)}{\sum_{i}\exp(\boldsymbol{q} \cdot \boldsymbol{k}_{i} / \epsilon)}
  \label{eq:method_nce}
\end{equation}

\noindent where $\mathcal{L}_{cl}$ denotes the InfoNCE~\cite{oord2018representation} loss function, $k_i$ means a (positive or arbitrarily negative) key sample, and $\epsilon=0.07$ is the temperature~\cite{wu2018unsupervised}. 
Following the unsupervised tracking fashion~\cite{liu2022online,shuai2022id}, the positive and negative keys mainly come from two sources within a video, \ie pseudo-labeled historical frames and self-augmented frames. 

\lk{However, two issues occur: (1) the uncertainty reduces the accuracy of pseudo-tracklets and (2) the randomly augmented samples fail to learn the inter-frame consistency.} 
We argue the above issues are not independent. 
\lk{By leveraging the uncertainty in turn,} the accurate pseudo-tracklets can guide the qualified positive and negative augmentations.

To address these two issues, we propose an uncertainty-aware pseudo-tracklet labeling strategy in \cref{sec:method_uoap}, which integrates a verification-and-rectification mechanism into the tracklet generation. 
Then we propose a tracklet-guided augmentation strategy in \cref{sec:method_ada_aug}, bringing temporary information into spatial augmentation. The augmented samples simulate the objects' motion. A hierarchical uncertainty-based sampling strategy is proposed for hard sample mining. More details are described in the following section.

\subsection{Uncertainty-aware Tracklet-Labeling}
\label{sec:method_uoap}

Accurate pseudo tracklet is critical in \liuk{learning feature consistency}. 
However, without manual annotation, \lk{the aggravated uncertainty makes} the tracklet-labeling a huge challenge due to various interference factors, including similar appearance among objects, frequent object cross and occlusions, \etc. 
\lk{In fact, the uncertainty can also be leveraged to improve the pseudo-accuracy in turn.}
In this section, we propose an \textbf{U}ncertainty-aware \textbf{T}racklet-\textbf{L}abeling (\textbf{UTL}) strategy for better pseudo-tracklets.

Given an input video sequence $V = \{I^{1}, I^{2}, \cdots, I^{N}\}$, each frame $I^{t}$ is annotated with the bounding boxes $B^{t} = \{b_{1}^{t}, b_{2}^{t}, \cdots, b_{M^{t}}^{t}\}$ of $M^{t}$ objects in $t_{th}$ frame, where $b_{i}^{t} = (cx_{i}^{t}, cy_{i}^{t}, w_{i}^{t}, h_{i}^{t})$ is the center coordinate and shape of the $i_{th}$ object $o_{i}^{t}$. As shown in~\cref{fig:method_fmwk}, \mywork~generates accurate pseudo-tracklets in four main steps:

1) \textbf{Association}. For a certain object $o_{i}^{t}$ in frame $I^{t}$, the $\ell_2$-normalized representation $\boldsymbol{f}_{i}^{t}$ can be expressed as $\boldsymbol{f}_{i}^{t} = {\phi}(I^{t}, b_{i}^{t})$, 
where the embedding encoder is denoted as $\phi$.

To associate the objects in frame $I^{t}$ with the objects or trajectories in previous $I^{t \minus 1}$, a similarity matrix is constructed with their appearance embeddings:

\begin{equation}
  \boldsymbol{C} \in \mathbb{R}^{M^{t} \times M^{t \minus 1}}, \;
  c_{i,j} = {\boldsymbol{f}_{i}^{t}} \cdot  \boldsymbol{f}_{j}^{t \minus 1}
  \label{eq:method_matrix}
\end{equation}

\noindent where $c_{i,j}$ represents the cosine similarity between the $i_{th}$ object in frame $I^{t}$ and the $j_{th}$ object (or trajectory) in frame $I^{t \minus 1}$. Then the Hungarian algorithm~\cite{kuhn1955hungarian} is adopted to generate the identity association results.

2) \textbf{Verification}. However, the appearance representations are sometimes unreliable, especially in the unsupervised scenario. To solve this issue, an uncertainty metric is proposed to evaluate the association after the first stage.






Object association can be viewed as a multi-category classification problem.
And confidence-score has been proved efficient and effective in detecting mis-classified examples~\cite{hendrycks2016baseline}.
Inspired by this, we propose to detect the mis-associated objects through the similarity scores.

Given an object $o_{i}^{t}$ associated with $o_{j}^{t \minus 1}$ in the previous frame based on \cref{eq:method_matrix}, the association ($o_{i}^{t} \!\sim\! o_{j}^{t \minus 1}$) is unconvincing in two cases: 
1) the assigned similarity $c_{i,j}$ is relatively low (\eg, partial occlusion or motion blur) and 
2) there are other objects whose similarities are close to the assigned $c_{i,j}$ (\eg, similar appearance or indistinguishable embedding).
It can be formulated as:

\begin{equation}
  c_{i,j} < m_1; \quad c_{i,j_{2}} > c_{i,j} - m_2
  \label{eq:method_margin}
\end{equation}

\noindent 
where $m_1,m_2$ are constant margins. For simplicity, only the second-highest similarity with others ($c_{i,j_{2}}$) is considered.
In an ideal association, $c_{i,j}$ should be close to 1 and $c_{i,j_{2}}$ close to 0.
We thus estimate the association \lk{risk} as:

\begin{equation}
  \sigma_{i,j} = - \log c_{i,j} - \log \left( 1 - c_{i,j_{2}} \right)
  \label{eq:method_energy}
\end{equation}

Detailed derivation is shown in \mappendix{B}.
Combining with \cref{eq:method_margin} and \cref{eq:method_energy} , an adaptive threshold is proposed:

\begin{equation}
  \gamma_{i,j} =  -\log m_1 - \log \left( 1 + m_2 - c_{i,j} \right)
  \label{eq:method_border}
\end{equation}

\begin{figure}[tb]
  \centering
  \begin{subfigure}{0.48\linewidth}
    \includegraphics[width=\linewidth]{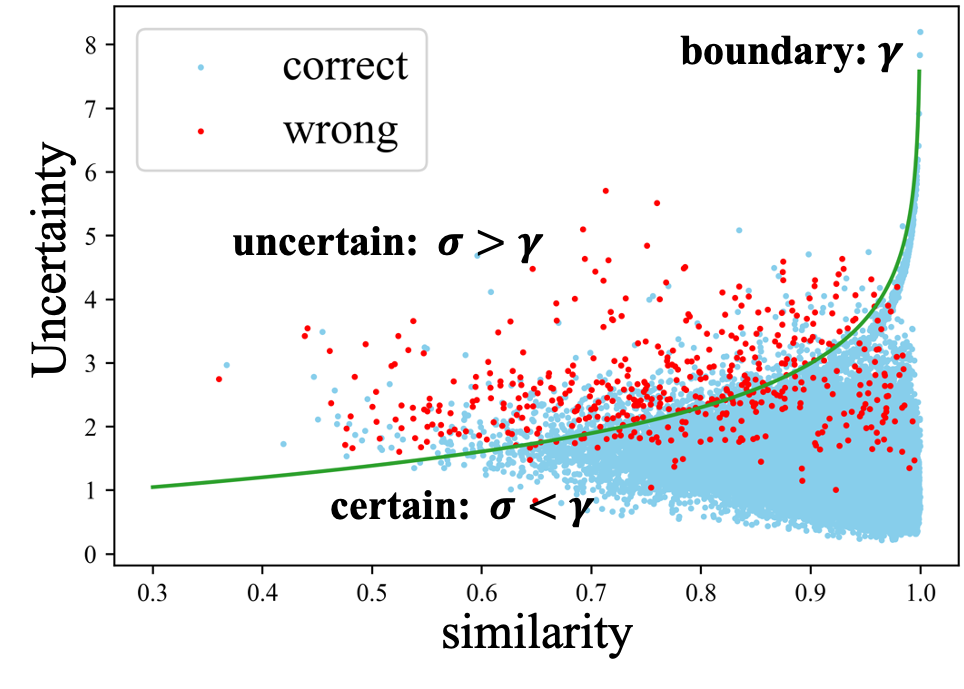}
    \caption{Association verification. }
    \label{fig:method_verify}
  \end{subfigure}
  \hfill
  \begin{subfigure}{0.50\linewidth}
    \includegraphics[width=\linewidth]{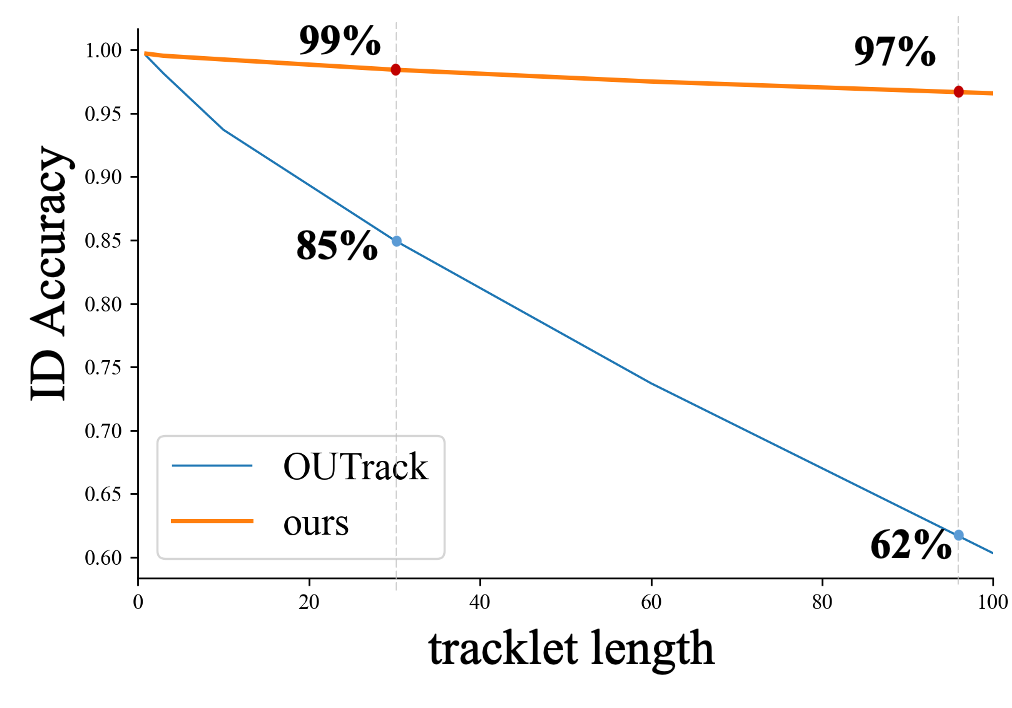}
    \caption{Pseudo Accuracy. }
    \label{fig:method_reidacc}
  \end{subfigure}
  \caption{\textbf{Statistics of pseudo-identities on MOT17-train set}. (a) 67\% wrong associations are divided into the `uncertain' area, and 99\% correct associations are preserved in `certain' area. (b) The pseudo-accuracy of previous methods dramatically drops as \lk{tracklet length} increases. By contrast, our \mywork~consistently maintains a high accuracy .}
  \label{fig:method_uncertain}
\end{figure}

As shown in~\cref{fig:method_verify}, when the \lk{risk} $\sigma_{i,j}$ is higher than the threshold $\gamma_{i,j}$, the assignment ($o_{i}^{t} \!\sim\! o_{j}^{t \minus 1}$) should be re-considered. 
\lk{The \textbf{association uncertainty} is quantified as:}

\begin{equation}
  \delta_{i,j} = \sigma_{i,j} - \gamma_{i,j}
  \label{eq:method_uncertain}
\end{equation}

The results are not sensitive to the exact margins. We set $m_1 = 0.5$ and $m_2 = 0.05$ for slightly better performance.

The uncertain pairs after the verification stage and unmatched objects after the association stage are gathered as uncertain candidates for the rectification stage.
We have provided several visualization examples to verify the certain/uncertain associations in \mappendix{I}.

3) \textbf{Rectification}. 
The rectification stage is performed among the uncertain candidate. The similarities between the two adjacent frames are no longer convincing.
More information should be taken into account, including motion \lk{estimation} and appearance \lk{variation} within a tracklet. 

For the uncertain candidates, \mywork~constructs another similarity matrix for the secondary rectification. 
First, \lk{the motion constraints should be relaxed}, so the association shares overlap \lk{higher than} $\beta$ 
\lk{are preserved}. 
Second, \lk{the appearance should not vary extremely fast}, so we adopt the averaged similarity between object $o_{i}^{t}$ and tracklet $trk_{j} = \{o_{j}^{t \minus K}, \cdots, o_{j}^{t \minus 1}\}$ within previous $K$ frames. 
In this stage, we solve the sub-problem of global identity assignments, which can be formulated as:

\begin{equation}
\begin{split}
  \boldsymbol{C}^\prime \in \mathbb{R}^{{M^{t}}^\prime \times {M^{t \minus 1}}^\prime} & \\
  c^\prime_{i,j} = \left( \frac{1}{K} \sum_{\hat{t} = t \minus K}^{t \minus 1} {\boldsymbol{f}_{i}^{t}} \cdot  \boldsymbol{f}_{j}^{\hat{t}} \right) 
            \times \mathbb{I} & \left( \text{IoU} \left( b_{i}^{t}, b_{j}^{t \minus 1} \right) > \beta \right) 
  \label{eq:method_recti}
\end{split}
\end{equation}

\noindent where $\mathbb{I}(*)$ is the indicator function. Then the match set is updated based on the Hungarian algorithm.

\lk{
\textit{Remark.} Our core contribution is the uncertainty-based verification mechanism, rather than the specific rectification, which shall be adjusted in practice. Empirically we set $\beta=0.1$ and $K=5$.
}

4) \textbf{Propagation}. The pseudo-tracklets are propagated frame-by-frame. As shown in~\cref{fig:method_reidacc}, our strategy brings \lk{consistently} accurate pseudo-identities, \lk{\eg, reaching 97\% accuracy across 100 frames}.
The long-term intra-tracklet consistency is successfully maintained.

It is worth mentioning that the \lk{verification and rectification} stages can be naturally applied to the inference process to boost the performance, \lk{which does not conflict with existing association methods}.

\subsection{Tracklet-Guided Augmentation}
\label{sec:method_ada_aug}

\begin{figure}[tb]
  \centering
  \begin{subfigure}{0.32\linewidth}
    \includegraphics[width=\linewidth]{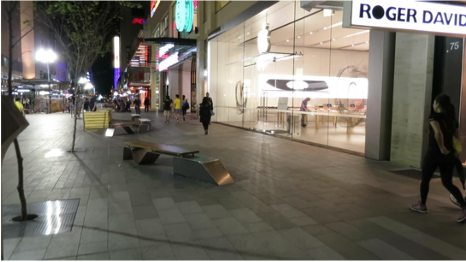}
    \caption{Current frame. }
    \label{fig:method_aug_cur_img}
  \end{subfigure}
  \hfill
  \begin{subfigure}{0.32\linewidth}
    \includegraphics[width=\linewidth]{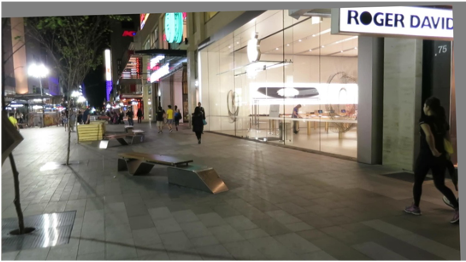}
    \caption{Weak RandAug. }
    \label{fig:method_aug_rand_aug_weak}
  \end{subfigure}
  \hfill
  \begin{subfigure}{0.32\linewidth}
    \includegraphics[width=\linewidth]{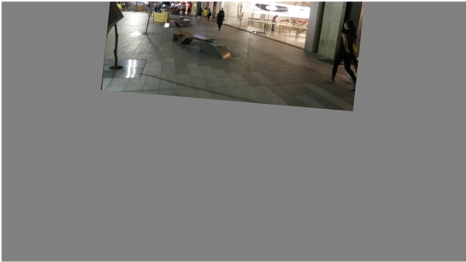}
    \caption{Strong RandAug. }
    \label{fig:method_aug_rand_aug_strong}
  \end{subfigure}
  \vfill
  \begin{subfigure}{0.32\linewidth}
    \includegraphics[width=\linewidth]{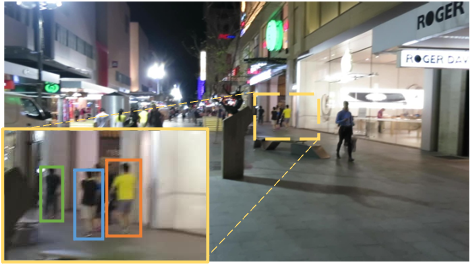}
    \caption{Historical frame. }
    \label{fig:method_aug_his_img}
  \end{subfigure}
  \hfill
  \begin{subfigure}{0.32\linewidth}
    \includegraphics[width=\linewidth]{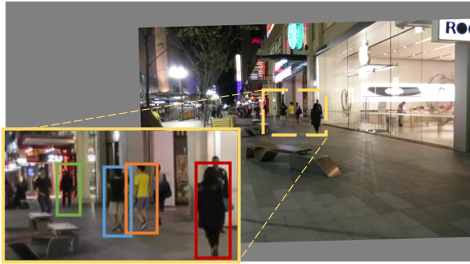}
    \caption{TGA: type-a. }
    \label{fig:method_aug_tga}
  \end{subfigure}
  \hfill
  \begin{subfigure}{0.32\linewidth}
    \includegraphics[width=\linewidth]{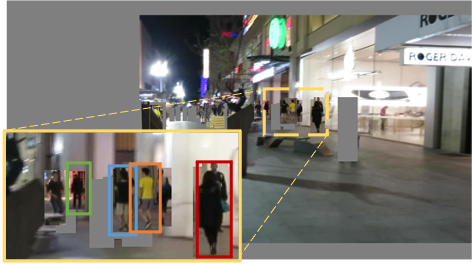}
    \caption{TGA: type-b. }
    \label{fig:method_aug_tga2}
  \end{subfigure}
  \caption{\textbf{Comparison of augmentations.} Random augmentations (\textbf{(b)} and \textbf{(c)}) fail to simulate the tracklet movements between frames (\textbf{(a)} and \textbf{(d)}), while TGA can. Two types of TGA are displayed: current objects + historical position + current \textbf{(e)} or historical \textbf{(f)} background.}
  \label{fig:method_ada_aug}
\end{figure}

Accurate pseudo-tracklets can guide the sample augmentation in the unsupervised MOT framework.
To learn the \liuk{inter-frame consistency}~\cite{chen2020simple,zhang2021fairmot}, good training samples should be diverse and \liuk{temporal-aware}. 
However, as illustrated in~\cref{fig:method_ada_aug}, existing methods usually treat augmentation and multi-object tracking as two isolated tasks, leading to ineffective augmentations. 
Instead, this paper utilizes the tracklet's spatial displacements to guide the augmentation process. 
Based on a properly selected anchor pair, the proposed strategy makes the augmented frames aligned to the historical frames, simulating realistic tracklet movements. The proposed method concurrently focuses on the hard negative samples.
Details of the \textbf{T}racklet-\textbf{G}uided \textbf{A}ugmentation (TGA) are given below.

Given a current frame $I^{t}$ with $M^{t}$ objects, a source-anchor object $o_{a}^{t}$ is selected, whose bounding box is denoted as $b_{a}^{t} = (cx_{a}^{t}, cy_{a}^{t}, w_{a}^{t}, h_{a}^{t})$. Then, we choose a target-anchor $o_{a}^{t \minus \tau}$ in $(t \minus \tau)_{th}$  historical frame from the pseudo-tracklet $trk_{a} = \{o_{a}^{t_0}, o_{a}^{t_1}, \cdots, o_{a}^{t}\}$. 
Finally, to augment the current $I^{t}$ to align with historical $I^{t \minus \tau}$,  a tracklet-guided affine transformation can be expressed as:

\begin{equation}
  \begin{bmatrix}
      x^{t \minus \tau} \\ y^{t \minus \tau} \\ 1
  \end{bmatrix}
  =
  \boldsymbol{M}_{t}^{t \minus \tau}
  \begin{bmatrix}
      x^{t} \\ y^{t} \\ 1
  \end{bmatrix}
  =
  \begin{bmatrix}
      m_{11} & m_{12} & m_{13} \\
      m_{21} & m_{22} & m_{23} \\
      0      & 0      & 1
  \end{bmatrix}
  \begin{bmatrix}
      x^{t} \\ y^{t} \\ 1
  \end{bmatrix}
  \label{eq:method_affine}
\end{equation}

\noindent where $x^*,y^*$ are spatial coordinates, and $\boldsymbol{M}_{t}^{t \minus \tau}$ can be solved by direct linear transform (DLT) algorithm ~\cite{detone2016deep}. 
Then an augmented frame $\tilde{I}^{t}$ is generated based on the tracklet-guided affine transformation with perspective jitter, which can be expressed as $\tilde{I}^{t} = \mathcal{T}\left(I^{t}, M_{t}^{t \minus \tau} \right)$.

Intuitively, a proper anchor-selection is vitally important for our augmentation strategy. 

First, the identity accuracy of anchor pair $(o_{a}^{t} \!\sim\! o_{a}^{t \minus \tau})$ is important. In other words, the consistency of anchor tracklet $trk_{a}$ should be guaranteed. We thus design a tracklet-level uncertain metric based on the propagated association-level uncertainty defined in \cref{eq:method_uncertain}, which is formulated as:

\begin{equation}
  \Omega_{i} = \frac{1}{n} \sum_{s=t_0}^{t} \exp (\delta_{i}^{s})
  \label{eq:method_tenergy}
\end{equation}

\noindent where $\Omega_{i}$ denotes the uncertainty of tracklet $trk_{i}$, and $n$ is the tracklet length.
An uncertainty-based sampling strategy is designed to select the source anchor $o_{a}^{t}$ (along with the anchor $trk_{a}$) from the $M^{t}$ objects in frame $I^{t}$ by:

\begin{equation}
  p\left(a=i \mid t \right) 
  = \frac{\exp{\left(-\Omega_{i}\right)}}{\sum_{\hat{i}=1}^{M^{t}}\exp{\left(-\Omega_{\hat{i}}\right)}}
  \label{eq:method_sel_an_src}
\end{equation}

\noindent where $p\left(a=i \mid t \right)$ represents the probability to choose the $i_{th}$ tracklet $trk_{i}$ as the anchor $trk_{a}$.
The uncertain tracklet with high $\Omega$ is less likely to be selected, avoiding dramatic augmentations from erroneous pseudo-tracklets.

Second, hard negative samples matter in discriminability learning. We tend to choose an indistinguishable (or, highly uncertain) target anchor $o_{a}^{t \minus \tau}$ along the tracklet $trk_{i}$. The selection probability can be formulated as:

\begin{equation}
  p\left(\pi=t \minus \tau \mid a \right) 
  = \frac{\exp{\left(\delta_{a}^{t \minus \tau}\right)}}{\sum_{\hat{\tau}=t_0}^{t-1}\exp{\left(\delta_{a}^{t-\hat{\tau}}\right)}}
  \label{eq:method_sel_an_tgt}
\end{equation}

Compared to conventional random transformation, our tracklet-guided augmentation is well-directed and tracking-related. 
A visualization example is displayed in the \mappendix{E} to illustrate the hierarchical sampling process.

\begin{figure}[htb]
  \setlength{\abovecaptionskip}{0.cm}
  \centering
  \includegraphics[width=0.81\linewidth]{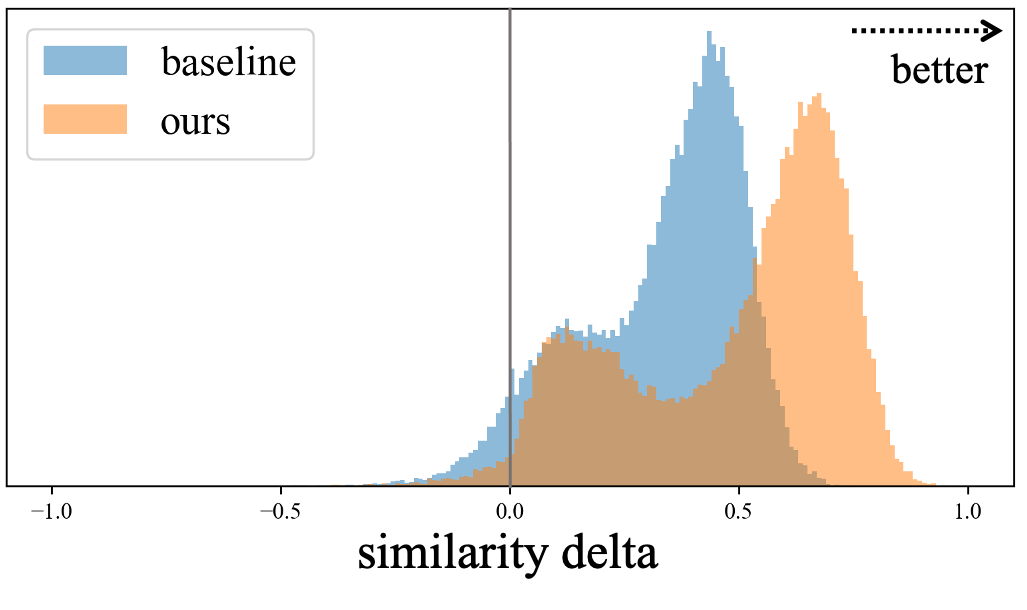}
  \caption{\lk{\textbf{Inter-frame consistency visualization.} During the associations, we statistic the similarity delta ($\Delta$) between ground-truth association ($c^{+}$) and other objects with the largest similarity ($c^{-}$). A positive delta ($\Delta = c^{+} - c^{-} > 0$) means a good tracker, the larger the better.} }
  \label{fig:method_disc_vis}
\end{figure}

Together with accurate pseudo-tracklets, the inter-frame consistency is successfully improved, as shown in \cref{fig:method_disc_vis}.

\section{Experiment}
\label{sec:exp}

\subsection{Datasets and Evaluation Metrics}
\label{sec:exp_setup}

\textbf{Datasets.}
 Experiments are performed on three popular benchmarks: MOT17~\cite{milan2016mot16}, MOT20~\cite{dendorfer2020mot20}, and the challenging VisDrone-MOT~\cite{zhu2021visdrone}.  MOT17 contains 14 videos captured from diverse viewpoints and weather conditions, while MOT20 consists of 8 videos on crowded scenes with heavier occlusion. Both of them are evaluated with the ``private detection'' protocol. 
VisDrone-MOT is captured by UAVs in various scenes, which comprises 79 sequences with 10 object categories. Only five categories (\ie, car, bus, truck, pedestrian, and van) are considered during evaluation~\cite{liu2022multi}. Multi-class tracking with irregular camera motion makes VisDrone-MOT a challenging benchmark.

\textbf{Metrics.} Following the previous MOT methods~\cite{li2022unsupervised,zhang2021fairmot,zhang2022bytetrack}, the HOTA~\cite{luiten2021hota} and CLEAR~\cite{bernardin2008evaluating} metrics are adopted to evaluate the trackers. Specifically, CLEAR mainly includes multiple object tracking accuracy (MOTA), ID F1 score (IDF1), and identity switches (IDS).

\subsection{Implementation Details}
\label{sec:exp_impl}

To show the efficacy of our unsupervised MOT framework, we implement~\mywork~\lk{on YOLOX~\cite{ge2021yolox}} with a ReID head integrated (see \mappendix{H}).
Specifically, the detection branch is trained in the same way as ByteTrack~\cite{zhang2022bytetrack}, while the new ReID head is learned with our \mywork. 
The model is trained with SGD optimizer and the initial learning rate of $1 \times 10^{-3}$ \lk{with cosine annealing schedule}.

For MOT17 and MOT20, the model is trained with the same setting as ByteTrack. Taking MOT20 for example, we train \mywork~for 80 epochs with an extra CrowdHuman~\cite{shao2018crowdhuman} dataset. 
For VisDrone-MOT, \mywork~is trained for 40 epochs without extra datasets. 
A pre-trained UniTrack~\cite{wang2021different} ReID model is added for ByteTrack to handle the multi-class multi-object tracking~\cite{zhang2022bytetrack}.
Specifically, the input image size is $1440 \times 800$ for MOT-challenges and $1600 \times 896$ for VisDrone-MOT.

\lk{Identity labels are unused in ALL training datasets.}

\begin{table}[tb]
    \centering
    \small
    \setlength{\tabcolsep}{2pt}
    \begin{tabular}{c|lc|cccc}
        \toprule
        Dataset & Tracker & \textit{Sup.} & HOTA$\uparrow$ & MOTA$\uparrow$ & IDF1$\uparrow$ & IDS$\downarrow$  \\  
        \midrule
        \multirow{12}{*}{MOT17} 
        & TrkFormer~\cite{meinhardt2022trackformer} & \CheckmarkBold & 57.3 & 74.1 & 68.0 & 2829 \\
        & MOTR~\cite{zeng2021motr} & \CheckmarkBold & 57.8 & 73.4 & 68.6 & 2439 \\
        & TraDeS~\cite{wu2021track} & \CheckmarkBold & 52.7 & 69.1 & 63.9 & 3555 \\
        & CorrTrack~\cite{wang2021multiple} & \CheckmarkBold & 60.7 & 76.5 & 73.6 & 3369 \\
        & MTrack~\cite{yu2022towards} & \CheckmarkBold & 60.5 & 72.1 & 73.5 & 2028 \\
        & OUTrack~\cite{liu2022online} & \XSolidBrush & 58.7 & 73.5 & 70.2 & 4122 \\
        & PointID~\cite{shuai2022id} & \XSolidBrush & $-$ & 74.2 & 72.4 & 2748 \\
        & UEANet~\cite{li2022unsupervised} & \XSolidBrush & 62.7 & 77.2 & 77.0 & \topb{1533} \\
        & ByteTrack~\cite{zhang2022bytetrack} & \XSolidBrush & \topb{63.1} & \topa{80.3} & \topb{77.3} & 2196 \\
        & \textbf{\mywork~}(Ours) & \XSolidBrush & \topa{64.2} & \topb{79.7} & \topa{78.2} & \topa{1506} \\
        \midrule
        \multirow{7}{*}{MOT20} 
        & CorrTrack~\cite{wang2021multiple} & \CheckmarkBold & $-$ & 65.2 & 69.1 & 5183 \\
        & MTrack~\cite{yu2022towards} & \CheckmarkBold & 55.3 & 63.5 & 69.2 & 6031 \\
        & OUTrack~\cite{liu2022online} & \XSolidBrush & 56.2 & 68.6 & 69.4 & 2223 \\
        & UEANet~\cite{li2022unsupervised} & \XSolidBrush & 58.6 & 73.0 & \topb{75.6} & 1423 \\
        & ByteTrack~\cite{zhang2022bytetrack} & \XSolidBrush & \topb{61.3} & \topa{77.8} & 75.2 & \topa{1223} \\
        & \textbf{\mywork~}(Ours) & \XSolidBrush & \topa{62.7} & \topb{77.1} & \topa{76.2} & \topb{1379} \\
        \bottomrule
    \end{tabular}
    \caption{\textbf{Performance comparison against SOTA trackers on MOT-Challenge test sets.} 
    `$\uparrow$'/`$\downarrow$' indicates higher/lower values are better, respectively. 
    \topa{Bold} numbers are superior results. 
    }
    \label{tab:res_mot}
\end{table}

\begin{table}[t]
    \centering
    \small
    \setlength{\tabcolsep}{4pt}
    \begin{tabular}{lc|ccccc}
        \toprule
        Method & \textit{Sup.}  & MOTA$\uparrow$ & IDF1$\uparrow$ & IDS$\downarrow$ & FPS$\uparrow$ \\
        \midrule
        MOTR~\cite{zeng2021motr}  & \CheckmarkBold  & 22.8 & 41.4 & 959 & 7.5 \\
        TrkFormer~\cite{meinhardt2022trackformer} & \CheckmarkBold & 24.0 & 30.5 & 4840 & 7.4 \\
        UavMOT~\cite{liu2022multi}  & \CheckmarkBold & 36.1 & 51.0 & 2775 & 12.0 \\
        ByteTrack~\cite{zhang2022bytetrack} & \XSolidBrush & 52.3 & 68.3 & 1232 & 11.4 \\
        \textbf{\mywork~}(Ours) & \XSolidBrush & \topa{52.3} & \topa{69.0} & \topb{1052} & \topa{19.4} \\
        \bottomrule
    \end{tabular}
    \caption{\textbf{Performance on VisDrone-MOT test-dev set.}}
    \label{tab:res_mcmot}
\end{table}

\subsection{Main Results}
\label{sec:exp_result}
\textbf{MOT-Challanges}. 
Evaluated by the official server, the results on MOT17 and MOT20 benchmarks are illustrated in \cref{tab:res_mot}, which shows \mywork~beats all of the SOTA supervised and unsupervised methods \lk{on HOTA and IDF1 metrics}. Specifically, it outperforms the SOTA unsupervised UEANet~\cite{li2022unsupervised} by a large margin (\eg, 1.2\% HOTA on MOT17). 
With the assistance of the ReID head, \mywork~consistently performs better in terms of HOTA and IDF1 against ByteTrack~\cite{zhang2022bytetrack}.
\lk{However, the IDS increases on MOT20, which is mainly because the extracted feature embedding is naturally biased in such severe scenarios. Embedding-based unsupervised methods (including our \mywork) are inferior to occluded similarities, leading to the IDS increase. Some occlusion-aware optimizations~\cite{he2020guided,yan2021occluded} might alleviate this problem.
}
In addition, the MOTA of \mywork~is slightly decreased, which implies that the competition between detection and re-identification tasks should be further explored. Detailed discussions and experiments are provided in \mappendix{A2}.

In addition, \mywork~does not involve network structure evolution, so the performance gains brought by \mywork~is uncorrelated with those enhancement modules proposed by advanced trackers in~\cref{tab:res_mot}. Combining \mywork~with these methods would lead to even better tracking performance.

\textbf{VisDrone-MOT}.
For the videos captured in UAV views, the IoU information (or motion model) is unreliable due to the irregular camera motion. 
To deal with this issue, camera motion compensation~\cite{zhang2022bytetrack} and objects' positional relation~\cite{liu2022multi} are mainly adopted, which are effective but computationally expensive. This work provides another solution, \lk{\ie, using an uncertainty metric and relaxed motion constraints for refined association results, which is robust to large camera motion as well as low frame rate}.
As shown in \cref{tab:res_mcmot}, \mywork~ substantially outperforms all the comparison methods,
\lk{including ByteTrack, who shares the same detector with our~\mywork~while utilizes the pre-trained self-supervised UniTrack model~\cite{wang2021different} for ReID. Also without identity annotation, the IDF1 gains demonstrate our learned task-specific appearance embedding head beats the pre-trained model.}
Besides, the improved FPS mainly comes from \mywork's jointly-trained ReID head, rather than an extra ReID model that requires another thorough inference on the raw images.
The results demonstrate the effectiveness and efficiency of our proposed \mywork~tracker.
Typical tracking visualizations are provided in \mappendix{J} to demonstrate our superiority.

\subsection{Ablation Studies}
\label{sec:exp_ablation}

In this section, we conduct extensive ablation studies to elaborate on the effectiveness of the proposed approach. Following the previous methods~\cite{zhang2021fairmot,li2022unsupervised,wu2021track}, the first half of each video of the MOT17 training set is used for training, and the second half is for validation. All the models are trained for 30 epochs.
\lk{Beside the results below, we also conduct ablation studies by training and testing on separate videos with cross-validation. The conclusion is unchanged. For more details please refer to \mappendix{G}.}

\begin{table}[t]
    \centering
    \begin{tabular}{c|cccc}
        \toprule
        Method & HOTA$\uparrow$ & MOTA$\uparrow$ & IDF1$\uparrow$ & IDS$\downarrow$  \\
        \midrule
        baseline & 63.40 & 73.73 & 74.51 & 207 \\
        +LTD & 63.43 & 73.74 & 74.64 & 202 \\
        +UTL & 63.84 & 73.78 & 75.19 & 203 \\
        +TGA & \topa{64.08} & 73.79 & \topa{75.42} & \topa{197} \\
        \midrule
        \textit{supervised}     & \textit{63.96} & \textit{73.79} & \textit{75.32} & \textit{196} \\
        \bottomrule
    \end{tabular}
    \caption{\textbf{Evaluation of the proposed modules.} 
    }
    \label{tab:ablation_component}
\end{table}

\begin{figure}[tb]
  \setlength{\abovecaptionskip}{0.cm}
  \centering
  \includegraphics[width=1.0\linewidth]{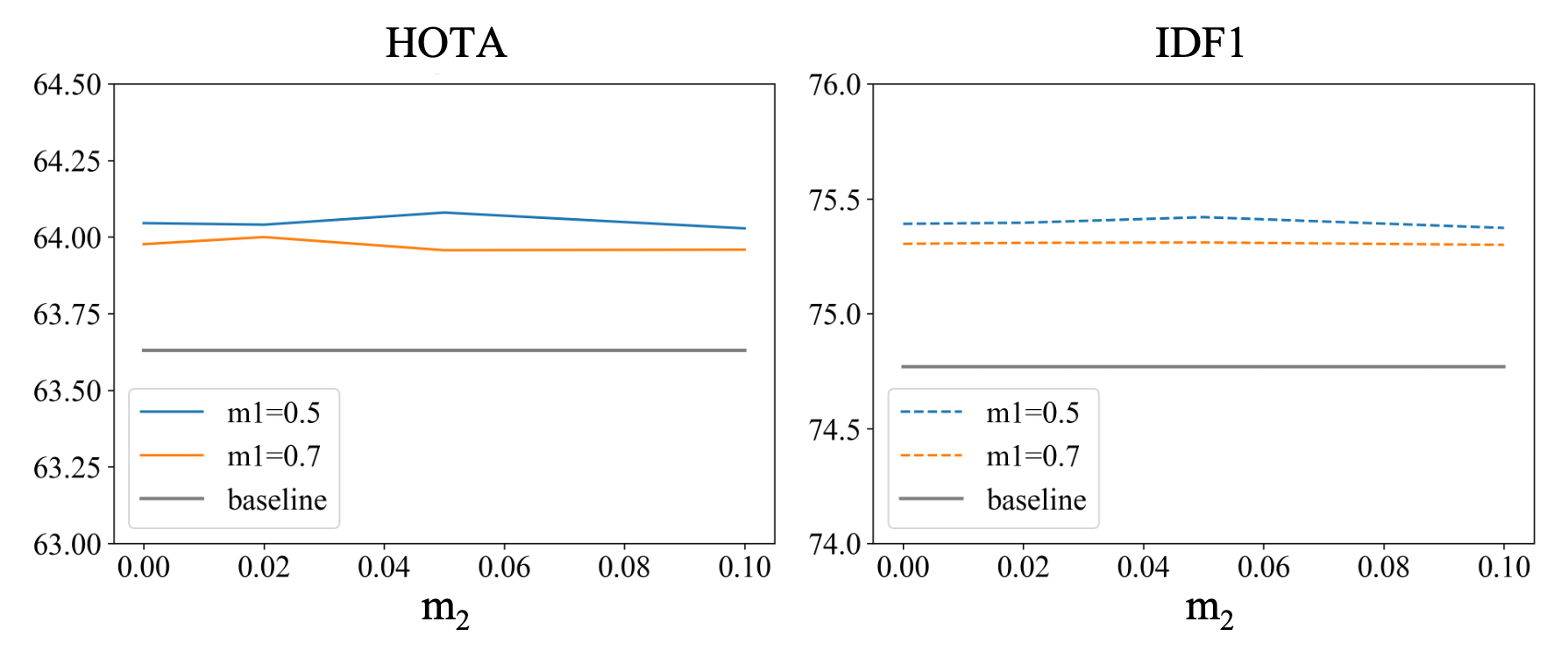}
  \caption{\lk{\textbf{Ablation on the proposed uncertainty metric.} } }
  \label{fig:abl_uncertain}
\end{figure}

\begin{table}[t]
    \centering
    \setlength{\tabcolsep}{4pt}
    \begin{tabular}{cc|cccc}
        \toprule
        TGA-src & TGA-tgt & HOTA$\uparrow$ & MOTA$\uparrow$ & IDF1$\uparrow$ & IDS$\downarrow$  \\
        \midrule
        $-$ & $-$    & 63.84 & 73.78 & 75.19 & 203 \\
        \midrule 
        random    & random    & 63.93 & 73.79 & 75.31 & 204 \\
        uncertain & random    & 63.92 & 73.77 & 75.36 & 194 \\
        random    & uncertain & 64.01 & 73.79 & 75.35 & 201 \\
        uncertain & uncertain & \topa{64.08} & 73.79 & \topa{75.42} & \topb{197} \\
        \bottomrule
    \end{tabular}
    \caption{\textbf{Ablation on the anchor-selection mechanism in TGA (\cref{sec:method_ada_aug}).} The ``$-$'' indicates TGA is not applied.
    }
    \label{tab:ablation_aug}
\end{table}

\textbf{Effectiveness of the modules proposed in \mywork.} 
Our unsupervised framework proposes two major components: uncertainty-aware tracklet-labeling (UTL) and tracklet-guided augmentation (TGA).
To evaluate each component, we conduct an ablation study on the tracking performance. The results are shown in \cref{tab:ablation_component}. 
We first construct a baseline model by training on adjacent frames. To introduce long-term dependency (LTD), the vanilla similarity-based association on historical frames is conducted for pseudo identities. However, it results in \lk{negligible gains in terms of HOTA and MOTA} due to the noisy pseudo-labels, \lk{meanwhile the IDF1 and IDS obtain slight increases}.
Instead, the proposed UTL strategy improves the tracking performance in \lk{most of the metrics (\eg, 0.4\% HOTA)}, which evidences the fact that long-term temporal consistency is well preserved. 
Finally, the TGA strategy results in increases of 0.2\% HOTA and 0.2\% IDF1, \lk{as well as decreased IDS}, demonstrating that our task-specific augmentation assists in learning the inter-frame consistency.
Equipped with the proposed components, unsupervised \mywork~even achieves better HOTA and IDF1 against the identity-supervised model (without UTL and TGA), which validates the effectiveness of our method and indicates the potential to leverage large-scale unlabeled videos.

\textbf{Uncertain margins.} 
Since the uncertainty metric is vital, we investigate the performance variance caused by different uncertainty margins when verifying the associations. As shown in \cref{fig:abl_uncertain}, different combinations of $m_1,m_2$ consistently improve the tracking performance. And the improvement is relatively not sensitive to the exact value of these two hyper-parameters. 
It indicates that wrong associations usually occur in candidates with comparable similarities and relatively lower confidence, which are able to be filtered out and rectified.
We choose $m_1=0.5,m_2=0.05$ for slightly better performance.
Moreover, we have provided further experiments on parameter stability with different models in \mappendix{D}, as well as a comparison with other uncertainty metrics in \mappendix{C} and \mappendix{F}.

\textbf{Augmentation strategies.} 
The customized tracklet-guided augmentation is mainly explored in \cref{tab:ablation_aug}, where the hierarchical uncertainty-based anchor-sampling mechanism is further evaluated.
First, TGA benefits the tracking performance even with totally random anchor-selections.
Meanwhile, the selected source anchor tracklet with low-uncertainty avoids dramatic transformation, which brings a slight decrease in IDS. Since most of the pseudo-tracklets are accurate enough after the training, this mechanism mostly serves as stabilizing the training in the early stages.
Moreover, the selected target anchor object with high-uncertainty along the tracklet brings qualified hard negative examples, leading to a 0.1\% HOTA increase.
Ultimately, the combined hierarchical uncertainty-based anchor-sampling mechanism results in better performance, demonstrating the effectiveness of TGA.
Furthermore, we have quantitatively evaluated the superiority of the TGA strategy over other approaches in \mappendix{F}.

\begin{table}[t]
    \centering
    \begin{tabular}{c|cccc}
        \toprule
        Tracker & HOTA$\uparrow$ & MOTA$\uparrow$ & IDF1$\uparrow$ & IDS$\downarrow$  \\
        \midrule
        \textbf{\mywork}   & 64.08 & 73.79 & 75.42 & \topa{197} \\
        \textbf{+UTL}       & \topa{64.90} & \topa{74.09} & \topa{76.66} & 212 \\
        \midrule
        ByteTrack  & 63.32 & 73.72 & 74.32 & \topa{207} \\
        \textbf{+UTL}       & \topa{64.90} & \topa{74.09} & \topa{76.66} & 212 \\
        \midrule
        FairMOT    & 62.03 & 72.65 & 72.83 & 618 \\
        \textbf{+UTL}       & \topa{64.01} & \topa{73.35} & \topa{75.05} & \topa{427} \\
        \midrule
        DeepSORT   & 58.48 & 70.81 & 66.20 & 526 \\
        \textbf{+UTL}       & \topa{59.75} & \topa{70.97} & \topa{67.60} & \topa{512} \\
        \midrule
        MOTDT      & 60.49 & 71.95 & 69.87 & 622 \\
        \textbf{+UTL}       & \topa{61.46} & \topa{72.55} & \topa{71.40} & \topa{353} \\
        \bottomrule
    \end{tabular}
    \caption{\textbf{Inference boosting.} Results are obtained by different association strategies with the SAME model.}
    \label{tab:ablation_infer_boost}
\end{table}

\textbf{Inference boosting.} The proposed uncertainty-aware tracklet labeling (UTL) strategy does not conflict with existing matching strategies. On the contrary, combined with our method, existing methods achieve better tracking performance. As shown in ~\cref{tab:ablation_infer_boost}, \lk{we first set our method as the comparison baseline, which simply adds ReID embeddings to the ByteTrack~\cite{zhang2022bytetrack}, and our UTL can thus be equipped. Besides, we integrate UTL to other three popular MOT trackers, including FairMOT~\cite{zhang2021fairmot}, DeepSORT~\cite{wojke2017simple}, and MOTDT~\cite{chen2018real}. It shows that the UTL consistently boosts all of these trackers by a large margin on most of the metrics, especially in HOTA and IDF1. The IDS of FairMOT and MOTDT is significantly decreased. The training-free UTL shows its effective and generalized application prospects. }

\begin{figure}[t]
  \centering
  \begin{subfigure}{1.0\linewidth}
    \includegraphics[width=\linewidth]{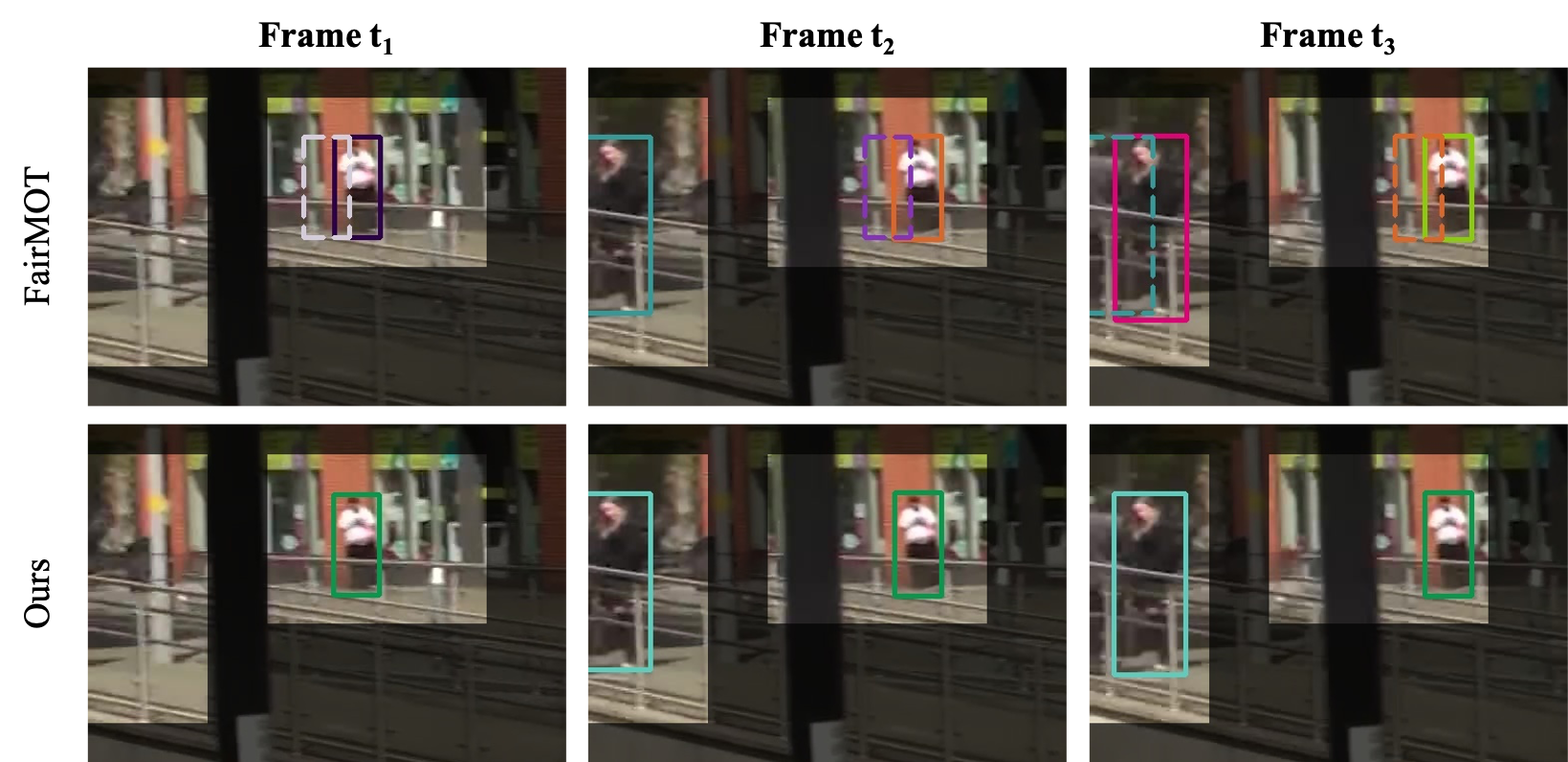}
    \caption{Fast Camera Motion. }
    \label{fig:vis_cam_move}
  \end{subfigure}
  \vfill
  \vfill
  \begin{subfigure}{1.0\linewidth}
    \includegraphics[width=\linewidth]{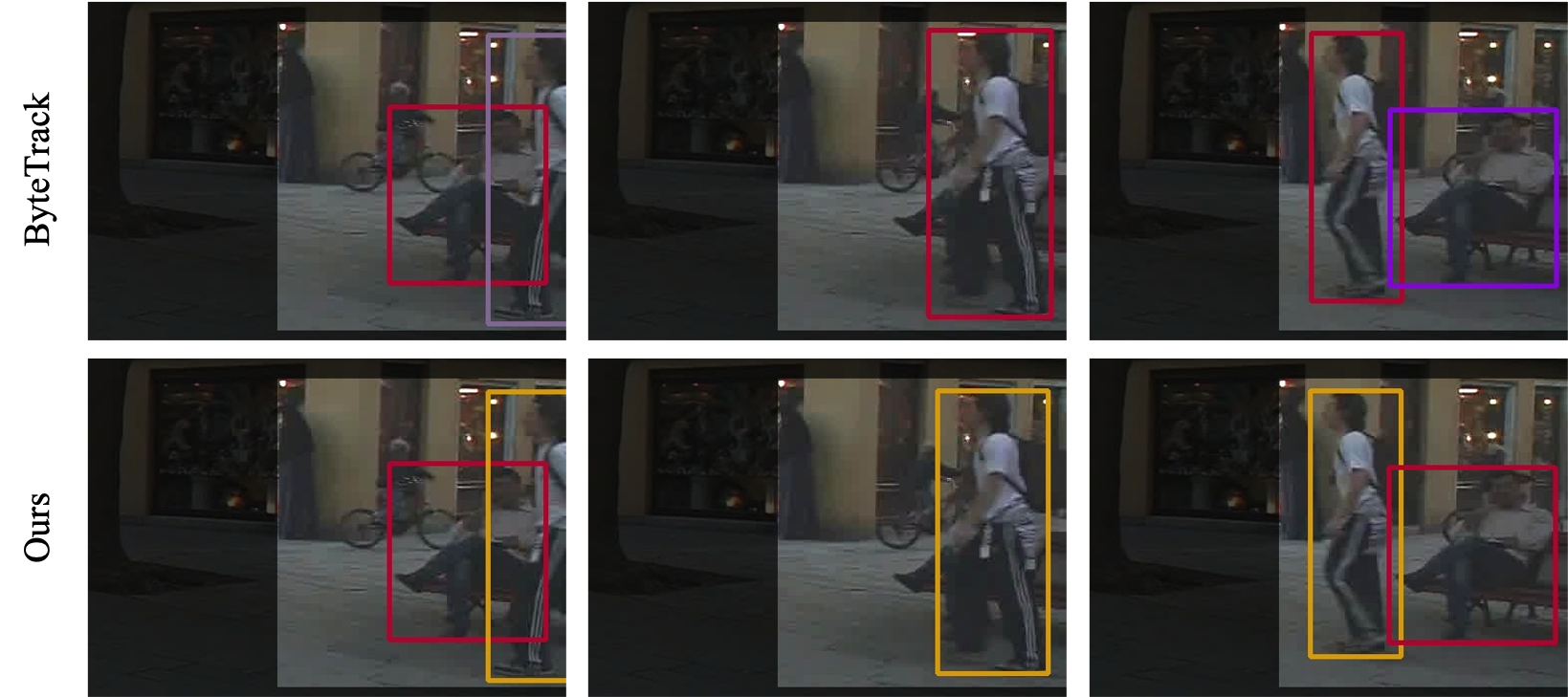}
    \caption{Object Occlusion. }
    \label{fig:vis_occ}
  \end{subfigure}
  \caption{\textbf{Typical visualizations.} (a) Under a moving camera, our method gets rid of the wrong motion prediction from Kalman Filter (dashed boxes). (b) In an occlusion case, tracklet features assist in avoiding ID switches.}
  \label{fig:visualize}
\end{figure}

Some typical visualization results are shown in \cref{fig:visualize}, which is consistent with ~\cref{tab:ablation_infer_boost}. 
First, when IoU information is unreliable in irregular camera motions, our method is robust to spatial prediction noise with the uncertainty-based verification stage. 
Second, in the rectification stage, the tracklet appearance embedding provides important supplementary information to confront the transient occlusions.

\section{Methodology Limitation}
\label{sec:lim}
While \mywork~can enhance the ability of unsupervised trackers by leveraging uncertainty during training, the current implementation has some limitations. One of these limitations is that the uncertainty assessment is conducted offline, which is isolated from the network training process. This means that the model cannot adjust and improve in real-time during training based on the uncertainty analysis, potentially limiting its ability to optimize its performance. Moreover, this offline uncertainty assessment has led to an increase in train time, with the current implementation taking twice as long to train the network. This could be problematic in scenarios where time is a critical factor or when there are large amounts of data to process.

\section{Conclusion}
\label{sec:conclu}
This paper proposes a novel unsupervised MOT framework, \mywork, to address the challenging and underexplored issue of uncertainty in visual tracking. The proposed method improves the quality of pseudo-tracklets through an uncertain-aware tracklet-labeling strategy and enhances tracklet consistency through a tracklets-guide augmentation method that employs a hierarchical uncertainty-based sampling approach for generating hard samples. Experimental results demonstrate the effectiveness of \mywork~,  showing the potential of uncertainty. Moving forwards, we will continue to investigate a general video-related uncertainty metric and its applications in various downstream tasks.

{\small
\bibliographystyle{ieee_fullname}
\bibliography{egbib}

\begin{thebibliography}{10}\itemsep=-1pt

\bibitem{bai2022directional}
Yalong Bai, Yifan Yang, Wei Zhang, and Tao Mei.
\newblock Directional self-supervised learning for heavy image augmentations.
\newblock In {\em Proceedings of the IEEE Conference on CVPR}, pages
  16692--16701, 2022.

\bibitem{bernardin2008evaluating}
Keni Bernardin and Rainer Stiefelhagen.
\newblock Evaluating multiple object tracking performance: the clear mot
  metrics.
\newblock {\em EURASIP Journal on Image and Video Processing}, 2008:1--10,
  2008.

\bibitem{bescos2021dynaslam}
Berta Bescos, Carlos Campos, Juan~D Tard{\'o}s, and Jos{\'e} Neira.
\newblock Dynaslam ii: Tightly-coupled multi-object tracking and slam.
\newblock {\em IEEE robotics and automation letters}, 6(3):5191--5198, 2021.

\bibitem{bewley2016simple}
Alex Bewley, Zongyuan Ge, Lionel Ott, Fabio Ramos, and Ben Upcroft.
\newblock Simple online and realtime tracking.
\newblock In {\em 2016 IEEE ICIP}, pages 3464--3468. IEEE, 2016.

\bibitem{chen2018real}
Long Chen, Haizhou Ai, Zijie Zhuang, and Chong Shang.
\newblock Real-time multiple people tracking with deeply learned candidate
  selection and person re-identification.
\newblock In {\em 2018 IEEE ICME}, pages 1--6. IEEE, 2018.

\bibitem{chen2020simple}
Ting Chen, Simon Kornblith, Mohammad Norouzi, and Geoffrey Hinton.
\newblock A simple framework for contrastive learning of visual
  representations.
\newblock {\em arXiv preprint arXiv:2002.05709}, 2020.

\bibitem{dendorfer2020mot20}
Patrick Dendorfer, Hamid Rezatofighi, Anton Milan, Javen Shi, Daniel Cremers,
  Ian Reid, Stefan Roth, Konrad Schindler, and Laura Leal-Taix{\'e}.
\newblock Mot20: A benchmark for multi object tracking in crowded scenes.
\newblock {\em arXiv preprint arXiv:2003.09003}, 2020.

\bibitem{detone2016deep}
Daniel DeTone, Tomasz Malisiewicz, and Andrew Rabinovich.
\newblock Deep image homography estimation.
\newblock {\em arXiv preprint arXiv:1606.03798}, 2016.

\bibitem{devries2018learning}
Terrance DeVries and Graham~W Taylor.
\newblock Learning confidence for out-of-distribution detection in neural
  networks.
\newblock {\em arXiv preprint arXiv:1802.04865}, 2018.

\bibitem{fan2018unsupervised}
Hehe Fan, Liang Zheng, Chenggang Yan, and Yi Yang.
\newblock Unsupervised person re-identification: Clustering and fine-tuning.
\newblock {\em ACM Transactions on Multimedia Computing, Communications, and
  Applications}, 14(4):1--18, 2018.

\bibitem{fawzi2016adaptive}
Alhussein Fawzi, Horst Samulowitz, Deepak Turaga, and Pascal Frossard.
\newblock Adaptive data augmentation for image classification.
\newblock In {\em 2016 IEEE international conference on image processing
  (ICIP)}, pages 3688--3692. Ieee, 2016.

\bibitem{ge2021yolox}
Zheng Ge, Songtao Liu, Feng Wang, Zeming Li, and Jian Sun.
\newblock Yolox: Exceeding yolo series in 2021.
\newblock {\em arXiv preprint arXiv:2107.08430}, 2021.

\bibitem{geiger2012we}
Andreas Geiger, Philip Lenz, and Raquel Urtasun.
\newblock Are we ready for autonomous driving? the kitti vision benchmark
  suite.
\newblock In {\em 2012 IEEE conference on CVPR}, pages 3354--3361. IEEE, 2012.

\bibitem{ghiasi2021simple}
Golnaz Ghiasi, Yin Cui, Aravind Srinivas, Rui Qian, Tsung-Yi Lin, Ekin~D Cubuk,
  Quoc~V Le, and Barret Zoph.
\newblock Simple copy-paste is a strong data augmentation method for instance
  segmentation.
\newblock In {\em Proceedings of the IEEE/CVF CVPR}, pages 2918--2928, 2021.

\bibitem{he2020momentum}
Kaiming He, Haoqi Fan, Yuxin Wu, Saining Xie, and Ross Girshick.
\newblock Momentum contrast for unsupervised visual representation learning.
\newblock In {\em Proceedings of the IEEE/CVF conference on CVPR}, pages
  9729--9738, 2020.

\bibitem{he2020guided}
Lingxiao He and Wu Liu.
\newblock Guided saliency feature learning for person re-identification in
  crowded scenes.
\newblock In {\em Computer Vision--ECCV 2020: 16th European Conference,
  Glasgow, UK, August 23--28, 2020, Proceedings, Part XXVIII 16}, pages
  357--373. Springer, 2020.

\bibitem{hendrycks2016baseline}
Dan Hendrycks and Kevin Gimpel.
\newblock A baseline for detecting misclassified and out-of-distribution
  examples in neural networks.
\newblock {\em International Conference on Learning Representations}, 2017.

\bibitem{jiang2021exploring}
Yiqi Jiang, Weihua Chen, Xiuyu Sun, Xiaoyu Shi, Fan Wang, and Hao Li.
\newblock Exploring the quality of gan generated images for person
  re-identification.
\newblock In {\em Proceedings of the 29th ACM International Conference on
  Multimedia}, pages 4146--4155, 2021.

\bibitem{karthik2020simple}
Shyamgopal Karthik, Ameya Prabhu, and Vineet Gandhi.
\newblock Simple unsupervised multi-object tracking.
\newblock {\em arXiv preprint arXiv:2006.02609}, 2020.

\bibitem{kuhn1955hungarian}
Harold~W Kuhn.
\newblock The hungarian method for the assignment problem.
\newblock {\em Naval research logistics quarterly}, 2(1-2):83--97, 1955.

\bibitem{lakshminarayanan2017simple}
Balaji Lakshminarayanan, Alexander Pritzel, and Charles Blundell.
\newblock Simple and scalable predictive uncertainty estimation using deep
  ensembles.
\newblock {\em Advances in Neural Information Processing Systems}, 30, 2017.

\bibitem{li2022unsupervised}
Yu-Lei Li.
\newblock Unsupervised embedding and association network for multi-object
  tracking.
\newblock In {\em Proceedings of the 31th International Joint Conference on
  Artificial Intelligence, (IJCAI-22)}, 2022.

\bibitem{lin2019bottom}
Yutian Lin, Xuanyi Dong, Liang Zheng, Yan Yan, and Yi Yang.
\newblock A bottom-up clustering approach to unsupervised person
  re-identification.
\newblock In {\em Proceedings of the AAAI conference on artificial
  intelligence}, pages 8738--8745, 2019.

\bibitem{liu2022online}
Qiankun Liu, Dongdong Chen, Qi Chu, Lu Yuan, Lei Zhang, and Nenghai Yu.
\newblock Online multi-object tracking with unsupervised re-identification
  learning and occlusion estimation.
\newblock {\em Neurocomputing}, 483:333--347, 2022.

\bibitem{liu2022multi}
Shuai Liu, Xin Li, Huchuan Lu, and You He.
\newblock Multi-object tracking meets moving uav.
\newblock In {\em Proceedings of the IEEE/CVF Conference on Computer Vision and
  Pattern Recognition}, pages 8876--8885, 2022.

\bibitem{liu2020energy}
Weitang Liu, Xiaoyun Wang, John Owens, and Yixuan Li.
\newblock Energy-based out-of-distribution detection.
\newblock {\em NeurIPS}, 33:21464--21475, 2020.

\bibitem{liu2021divaug}
Zirui Liu, Haifeng Jin, Ting-Hsiang Wang, Kaixiong Zhou, and Xia Hu.
\newblock Divaug: Plug-in automated data augmentation with explicit diversity
  maximization.
\newblock In {\em Proceedings of the IEEE/CVF International Conference on
  Computer Vision}, pages 4762--4770, 2021.

\bibitem{luiten2021hota}
Jonathon Luiten, Aljosa Osep, Patrick Dendorfer, Philip Torr, Andreas Geiger,
  Laura Leal-Taix{\'e}, and Bastian Leibe.
\newblock Hota: A higher order metric for evaluating multi-object tracking.
\newblock {\em International journal of computer vision}, 129(2):548--578,
  2021.

\bibitem{malinin2018predictive}
Andrey Malinin and Mark Gales.
\newblock Predictive uncertainty estimation via prior networks.
\newblock {\em Advances in Neural Information Processing Systems}, 31, 2018.

\bibitem{meinhardt2022trackformer}
Tim Meinhardt, Alexander Kirillov, Laura Leal-Taixe, and Christoph
  Feichtenhofer.
\newblock Trackformer: Multi-object tracking with transformers.
\newblock In {\em Proceedings of the IEEE/CVF Conference on Computer Vision and
  Pattern Recognition}, pages 8844--8854, 2022.

\bibitem{milan2016mot16}
Anton Milan, Laura Leal-Taix{\'e}, Ian Reid, Stefan Roth, and Konrad Schindler.
\newblock Mot16: A benchmark for multi-object tracking.
\newblock {\em arXiv preprint arXiv:1603.00831}, 2016.

\bibitem{oh2011large}
Sangmin Oh, Anthony Hoogs, Amitha Perera, Naresh Cuntoor, Chia-Chih Chen,
  Jong~Taek Lee, Saurajit Mukherjee, JK Aggarwal, Hyungtae Lee, Larry Davis,
  et~al.
\newblock A large-scale benchmark dataset for event recognition in surveillance
  video.
\newblock In {\em CVPR 2011}, pages 3153--3160. IEEE, 2011.

\bibitem{oord2018representation}
Aaron van~den Oord, Yazhe Li, and Oriol Vinyals.
\newblock Representation learning with contrastive predictive coding.
\newblock {\em arXiv preprint arXiv:1807.03748}, 2018.

\bibitem{said2012real}
Tarek Said, Samy Ghoniemy, and Omar Karam.
\newblock Real-time multi-object detection and tracking for autonomous robots
  in uncontrolled environments.
\newblock In {\em 2012 Seventh ICCES}, pages 67--72. IEEE, 2012.

\bibitem{sensoy2018evidential}
Murat Sensoy, Lance Kaplan, and Melih Kandemir.
\newblock Evidential deep learning to quantify classification uncertainty.
\newblock {\em NeurIPS}, 31, 2018.

\bibitem{shao2018crowdhuman}
Shuai Shao, Zijian Zhao, Boxun Li, Tete Xiao, Gang Yu, Xiangyu Zhang, and Jian
  Sun.
\newblock Crowdhuman: A benchmark for detecting human in a crowd.
\newblock {\em arXiv preprint arXiv:1805.00123}, 2018.

\bibitem{shuai2022id}
Bing Shuai, Xinyu Li, Kaustav Kundu, and Joseph Tighe.
\newblock Id-free person similarity learning.
\newblock In {\em Proceedings of the IEEE/CVF Conference on Computer Vision and
  Pattern Recognition}, pages 14689--14699, 2022.

\bibitem{sun2020scalability}
Pei Sun, Henrik Kretzschmar, Xerxes Dotiwalla, Aurelien Chouard, Vijaysai
  Patnaik, Paul Tsui, James Guo, Yin Zhou, Yuning Chai, Benjamin Caine, et~al.
\newblock Scalability in perception for autonomous driving: Waymo open dataset.
\newblock In {\em Proceedings of the IEEE/CVF conference on CVPR}, pages
  2446--2454, 2020.

\bibitem{wang2021can}
Haoran Wang, Weitang Liu, Alex Bocchieri, and Yixuan Li.
\newblock Can multi-label classification networks know what they don’t know?
\newblock {\em NeurIPS}, 34:29074--29087, 2021.

\bibitem{wang2019data}
Hao Wang, Qilong Wang, Fan Yang, Weiqi Zhang, and Wangmeng Zuo.
\newblock Data augmentation for object detection via progressive and selective
  instance-switching.
\newblock {\em arXiv preprint arXiv:1906.00358}, 2019.

\bibitem{wang2021multiple}
Qiang Wang, Yun Zheng, Pan Pan, and Yinghui Xu.
\newblock Multiple object tracking with correlation learning.
\newblock In {\em Proceedings of the IEEE/CVF Conference on Computer Vision and
  Pattern Recognition}, pages 3876--3886, 2021.

\bibitem{wang2019learning}
Xiaolong Wang, Allan Jabri, and Alexei~A Efros.
\newblock Learning correspondence from the cycle-consistency of time.
\newblock In {\em Proceedings of the IEEE/CVF Conference on Computer Vision and
  Pattern Recognition}, pages 2566--2576, 2019.

\bibitem{wang2020cycas}
Zhongdao Wang, Jingwei Zhang, Liang Zheng, Yixuan Liu, Yifan Sun, Yali Li, and
  Shengjin Wang.
\newblock Cycas: Self-supervised cycle association for learning re-identifiable
  descriptions.
\newblock In {\em European Conference on Computer Vision}, pages 72--88.
  Springer, 2020.

\bibitem{wang2021different}
Zhongdao Wang, Hengshuang Zhao, Ya-Li Li, Shengjin Wang, Philip Torr, and Luca
  Bertinetto.
\newblock Do different tracking tasks require different appearance models?
\newblock {\em NeurIPS}, 34:726--738, 2021.

\bibitem{wen2019batchensemble}
Yeming Wen, Dustin Tran, and Jimmy Ba.
\newblock Batchensemble: an alternative approach to efficient ensemble and
  lifelong learning.
\newblock In {\em International Conference on Learning Representations}, 2019.

\bibitem{wojke2017simple}
Nicolai Wojke, Alex Bewley, and Dietrich Paulus.
\newblock Simple online and realtime tracking with a deep association metric.
\newblock In {\em 2017 IEEE ICIP}, pages 3645--3649. IEEE, 2017.

\bibitem{wu2021track}
Jialian Wu, Jiale Cao, Liangchen Song, Yu Wang, Ming Yang, and Junsong Yuan.
\newblock Track to detect and segment: An online multi-object tracker.
\newblock In {\em Proceedings of the IEEE/CVF conference on Computer Vision and
  Pattern Recognition}, pages 12352--12361, 2021.

\bibitem{wu2018unsupervised}
Zhirong Wu, Yuanjun Xiong, Stella~X Yu, and Dahua Lin.
\newblock Unsupervised feature learning via non-parametric instance
  discrimination.
\newblock In {\em CVPR}, pages 3733--3742, 2018.

\bibitem{yan2021occluded}
Cheng Yan, Guansong Pang, Jile Jiao, Xiao Bai, Xuetao Feng, and Chunhua Shen.
\newblock Occluded person re-identification with single-scale global
  representations.
\newblock In {\em Proceedings of the IEEE/CVF International Conference on
  Computer Vision}, pages 11875--11884, 2021.

\bibitem{yu2022towards}
En Yu, Zhuoling Li, and Shoudong Han.
\newblock Towards discriminative representation: Multi-view trajectory
  contrastive learning for online multi-object tracking.
\newblock In {\em Proceedings of the IEEE/CVF Conference on CVPR}, pages
  8834--8843, 2022.

\bibitem{zeng2021motr}
Fangao Zeng, Bin Dong, Yuang Zhang, Tiancai Wang, Xiangyu Zhang, and Yichen
  Wei.
\newblock Motr: End-to-end multiple-object tracking with transformer.
\newblock In {\em European Conference on Computer Vision (ECCV)}, 2022.

\bibitem{zhan2021spatial}
Fangneng Zhan and Changgong Zhang.
\newblock Spatial-aware gan for unsupervised person re-identification.
\newblock In {\em 25th ICPR}, pages 6889--6896. IEEE, 2021.

\bibitem{zhang2022rethinking}
Junbo Zhang and Kaisheng Ma.
\newblock Rethinking the augmentation module in contrastive learning: Learning
  hierarchical augmentation invariance with expanded views.
\newblock In {\em Proceedings of the IEEE/CVF Conference on CVPR}, pages
  16650--16659, 2022.

\bibitem{zhang2022bytetrack}
Yifu Zhang, Peize Sun, Yi Jiang, Dongdong Yu, Fucheng Weng, Zehuan Yuan, Ping
  Luo, Wenyu Liu, and Xinggang Wang.
\newblock Bytetrack: Multi-object tracking by associating every detection box.
\newblock In {\em Proceedings of the European Conference on Computer Vision
  (ECCV)}, 2022.

\bibitem{zhang2021fairmot}
Yifu Zhang, Chunyu Wang, Xinggang Wang, Wenjun Zeng, and Wenyu Liu.
\newblock Fairmot: On the fairness of detection and re-identification in
  multiple object tracking.
\newblock {\em International Journal of Computer Vision}, 129(11):3069--3087,
  2021.

\bibitem{zhu2018vision}
Pengfei Zhu, Longyin Wen, Xiao Bian, Haibin Ling, and Qinghua Hu.
\newblock Vision meets drones: A challenge.
\newblock {\em arXiv preprint arXiv:1804.07437}, 2018.

\bibitem{zhu2021visdrone}
Pengfei Zhu, Longyin Wen, Dawei Du, Xiao Bian, Heng Fan, Qinghua Hu, and Haibin
  Ling.
\newblock Detection and tracking meet drones challenge.
\newblock {\em IEEE Transactions on Pattern Analysis and Machine Intelligence},
  pages 1--1, 2021.

\end{thebibliography}
}

\end{document}